\documentclass[runningheads]{llncs}
\usepackage{graphicx}

\usepackage[textwidth=15\marginparwidth,
 backgroundcolor=lightgray,
 linecolor=orange,
 textsize=scriptsize]{todonotes}
\usepackage{microtype}

\begin{document}
\title{Curriculum Learning in Deep Neural Networks for Financial Forecasting \thanks{Supported by Microsoft Corp., where all research was conducted.}}
%
%
\author{Allison Koenecke\inst{1,2} \orcidID{0000-0002-6233-8256} \and
Amita Gajewar\inst{2}
}
\authorrunning{A. Koenecke and A. Gajewar}
%
\institute{Stanford University, Stanford CA 94305, USA \\
\email{koenecke@stanford.edu} \\
\and
Microsoft Corp., Sunnyvale CA 94089, USA \\
\email{amitag@microsoft.com} \\
}
\maketitle              
\begin{abstract}
For any financial organization, computing accurate quarterly forecasts for various products is one of the most critical operations. As the granularity at which forecasts are needed increases, traditional statistical time series models may not scale well. We apply deep neural networks in the forecasting domain by experimenting with techniques from Natural Language Processing (Encoder-Decoder LSTMs) and Computer Vision (Dilated CNNs), as well as incorporating transfer learning. A novel contribution of this paper is the application of curriculum learning to neural network models built for time series forecasting. We illustrate the performance of our models using Microsoft's revenue data corresponding to Enterprise, and Small, Medium \& Corporate products, spanning approximately 60 regions across the globe for 8 different business segments, and totaling in the order of tens of billions of USD.  We compare our models' performance to the ensemble model (of traditional statistics and machine learning) currently being used by Microsoft Finance. Using this in-production model as a baseline, our experiments yield an approximately 30\% improvement overall in accuracy on test data. We find that our curriculum learning LSTM-based model performs best, which shows that one can implement our proposed methods without overfitting on medium-sized data.

\end{abstract}
\section{Introduction}
\label{sec:intro}
A key aspect of effective business planning is the ability to accurately forecast finances.  This paper is the result of a partnership with Microsoft's Finance team to provide them guidance on projected revenue for both their Enterprise, and Small, Medium \& Corporate (SMC) Groups. 

Our goal is to forecast the revenue for Microsoft products, wherein the world-wide revenue is partitioned into 8 different segments; examples of segments include Commercial Enterprise or SMC Education. Each segment is further partitioned into approximately 60 regions, and each region's revenue is then partitioned further into 20 different products.  We henceforth refer to each combination of segment, region, and product as a ``datarow." Overall, there are approximately 6,000 datarows (since not all products are sold in all regions), with each datarow corresponding to a time series used for our forecasting problem.

The forecasting models currently used by the Finance team are built using traditional time series and machine learning models \cite{amita}. Here, we extend the capability to forecast product revenue at a more granular level, with improved accuracy and efficiency.  As data become more granular, further insights can be made by the sales team.  However, this comes with obvious challenges in training any machine learning model: the historical length of revenue information available may vary across sub-levels, and data itself can become noisy for different sub-levels. In these cases, fitting a statistical time series model to individual time series may not necessarily yield accurate forecasts, nor would this be a scalable solution. Following recent advances in applying deep neural networks (DNNs) in the time series domain \cite{deepar}, this paper explores applying Long Short-Term Memory (LSTM) and Dilated Convolutional Neural Network (DCNN) models to hierarchical financial time series. 


In the following sections, we describe the advancements made from prior work, our data structure, and the two overarching DNN models used (LSTM and DCNN).  Specifically, we describe the incremental accuracy gains from pre-processing techniques and additional features included in each DNN model, highlighting the performance of Curriculum Learning as described below.  Lastly, we interpret results and discuss implications and future steps.  By comparing against Microsoft's production baseline accuracy, we find that our curriculum learning method can be successfully applied to various neural networks on time series data to achieve higher accuracy and positive results in bias and variance.

\section{Related Work}

Sequence-to-sequence modeling for time series has been fairly popular for the past several years, not just in industry, but also broadly from classrooms \cite{lstm_tut} to Kaggle \cite{cnn_tut}.  These methods range from vanilla models to advanced industry competitors.

There are three major differentiating features between our research and previous related work on time series forecasting. First, curriculum learning (as defined below) has not yet been applied to time series trends.  Second, we highlight the transfer learning occurring within-task, from datarows having enough historical data to train effectively, to datarows lacking the amount of historical data needed to serve as model inputs.  Third, while deep learning models implemented in industry are mostly applied to ``big data", this paper shows that both Recurrent Neural Networks (RNNs) and Convolutional Neural Networks (CNNs) can be used effectively on medium-sized data without overfitting.

We first address the previous work on curriculum learning, which is essentially changing the order of inputs to a model to improve results. The intuition from Natural Language Processing (NLP) regarding this method is that shorter sentences are easier to learn than longer sentences; so, without initialization, one can bootstrap via iterated learning in order of increasing sentence length.  The relevant literature, including specifically the described Baby Steps algorithm \cite{babysteps,leapfrog}, has been applied to LSTMs for parsing a Wall Street Journal corpus \cite{leapfrog}, n-gram language modeling \cite{ngram}, and for performing digit sums using LSTMs \cite{babysums}.  However, there has been no application of this work to real numerical time series data.

We next comment on how we have utilized the concept of transfer learning in our work.  While there has been work done on transfer learning across tasks for CNNs \cite{transfer} and RNNs \cite{med}, as well as research on  meta-learning across time series \cite{meta}, there has not yet been an extensively applied example showing the ability to ameliorate the missing data problem by forecasting one datarow using historical trends from (in our case) a different region, segment, or product. Prior work on similar transfer learning focuses on robustness of out-of-sample test results and testing predictions at different timesteps \cite{rama}, which does not account for missing data and is relatively infeasible to reproduce given the much smaller size of the Microsoft data.  We discuss implications at length in Section \ref{sec:discussion}.

Lastly, we turn to previous instances of using neural networks to forecast time series data.  While it is fairly straightforward to use neural networks on large datasets, it is more difficult to apply these techniques to small and medium-sized data due to the risk of overfitting.  Many companies have adopted the use of LSTMs for time series modeling, but arguably the most advanced public methodology comes from Uber, which won the 2018 M4 Forecasting Competition using a hybrid Exponential Smoothing and RNN model \cite{uber}.  Their work shares many basic elements with our work: a rolling window train and validation method; data preprocessing methods that involve deseasonalization; and the use of LSTMs.  However, Uber's application is quite different: first, their data are orders of magnitude larger than ours, and second, their data do not contain similarly rigid hierarchical elements (rather, their vast number of covariates necessitates an autoencoder for feature extraction).  Another proven neural network method for financial forecasting is the Dilated CNN \cite{cnn}, wherein the underlying architecture comes from DeepMind's WaveNet project \cite{wavenet}.  This prior work is again on data much larger than ours, and also does not specify or discuss many data pre-processing steps (after audio pre-processing, WaveNet simply quantizes to a fixed range).  However, we have found that certain pre-processing techniques, such as log-transformation of de-meaned values and de-seasonalization, can be crucial to improving accuracy.

\section{Data}

\subsection{Data Structure}
\begin{figure}
\centering
\includegraphics[width=0.9\textwidth]{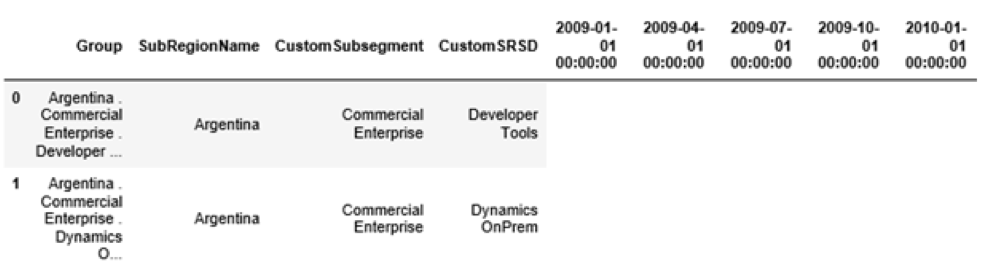}
\caption{Sample datarow structure excluding financial values} 
\label{fig:data}
\end{figure}

As noted in Section \ref{sec:intro}, world-wide revenues for Enterprise and SMC groups are partitioned into 8 business segments; each segment is partitioned into approximately 60 regions, and each region’s revenue is partitioned further into 20 different products.  Given historical quarterly revenue data, our goal is to forecast quarterly revenue for these products per combination of product, segment, and region; we then generate the aggregated segment-level forecasts as well as  world-wide aggregates.  Note that we focus on segment-level (rather than subregion or product-level) forecasts for comparison's sake, since this level has historically been used by the business.  All revenue numbers are adjusted to be in USD currency using a constant exchange rate.  Sample datarow structure is presented in Figure \ref{fig:data}.  

We use the quarterly revenue available (post-data-processing) for each datarow over fiscal quarters from January 2009 through July 2018 (totaling 39 timesteps).  Broadly speaking, we train on the first 35 timesteps of all datarows, and test on the final 4 timesteps; details are presented in the following section.

For all DNN models, we train on a subset of the data which has good enough history to fit a reasonable model.  Post-training, we apply this model to forecast revenue both for datarows on which it was trained, and also on out-of-sample datarows that were not seen by the model at the time of training due to insufficient historical information. Specifically, we perform basic data cleaning, and then use a subset of datarows (approximately 84\% of all datarows) containing sufficient history for model training.  We later apply transfer learning to the remaining out-of-sample datarows (approximately 16\% of all datarows).  Our results are evaluated by calculating Mean Absolute Percentage Error (MAPE) at the segment and world-wide level. 

\subsection{Microsoft Baseline}

Circa 2015, most of the revenue forecasting in Microsoft’s Finance division was driven by human judgement. In order to explore more efficient, accurate and unbiased revenue forecasting, machine learning methodology was explored along with statistical time series models \cite{amita2}. The methodology described in \cite{amita2} was used to compute forecasts in 13 worldwide regions. In \cite{amita}, this approach was further extended to use product level data available within each region, and to generate forecasts for each product within region (allowing for aggregation to region-level and world-wide forecasts).   This is the approach that is currently adopted by Microsoft’s Finance team; models based on this approach are running in a production environment to generate quarterly revenue forecasts to be used by Finance team members.  The results obtained from this method are referred to as the Microsoft baseline, and this paper explores whether the proposed DNN-based models can outperform the current baseline model in production.  For the ease of understanding the baseline model referred to in this paper, we describe here the methodology in \cite{amita} at a high level.

A product's historical revenue information varies depending on the age and popularity of the product; hence, it is not possible to naively apply single time series or machine learning model for all products, and still obtain accurate results. For very new products (having fewer than 6 quarters of revenue data), a simple heuristic is used. Otherwise, products are divided into three categories depending on the amount of historical revenue information available:

\begin{enumerate}
    \item{Products with more than 20 quarters of revenue data.  Microsoft uses a combined approach of various time series and machine learning models with cross validation for hyper-parameter tuning, where the final forecast generated corresponds to one of the time series (e.g., ARIMA, ETS and STL) or machine learning models (e.g., Random Forest, ElasticNet, etc.) that had the lowest historical error as computed on the validation dataset.}

    \item{Products with between 14 and 19 quarters of revenue data. Only statistical time series models are fit. Derived features are also constructed from these time series models, e.g., the average of the ETS-forecast and ARIMA-forecast can be used as an additional forecasted data point.}
    
    \item{Products with between 6 and 13 quarters of revenue data.  Only ARIMA and ETS statistical time series models are fit, as STL cannot be trained on very short time series. Since there is not enough history available to set aside a validation dataset, the final forecast is the simple average of the ARIMA-forecast and ETS-forecast. } 
\end{enumerate}

In aggregate, the above methods described form the Microsoft baseline that will be used as a benchmark for our results described below.

\section{Methods}

Our work on time series is mostly inspired by non-financial applications.  Specifically, Encoder-Decoder LSTMs (Section \ref{lstm}) are used in NLP, and Dilated CNNs (Section \ref{dcnn}) are applied in Computer Vision and Speech Recognition.

\subsection{RNN Model: Encoder-Decoder LSTM}
\label{lstm}

\begin{figure}[h]
  \centering
  \begin{minipage}[h]{0.48\textwidth}
    \includegraphics[width=0.98\textwidth]{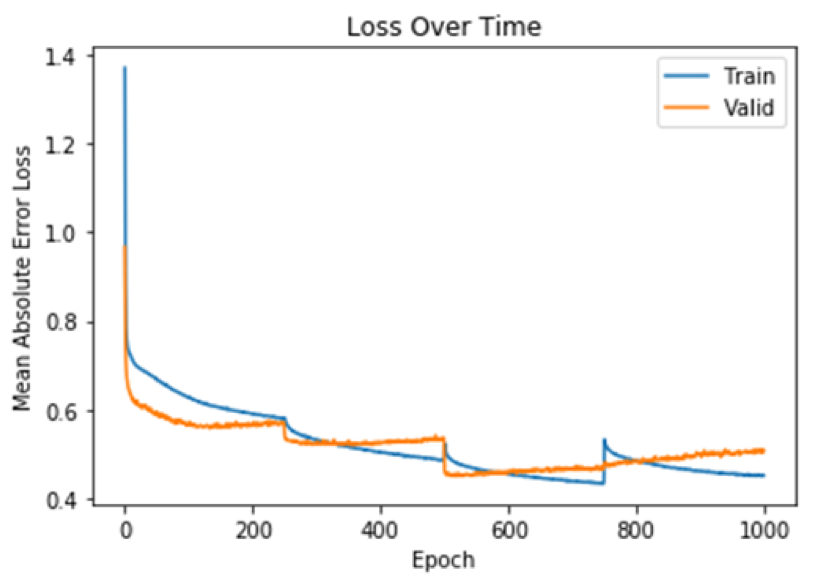}
    \caption{Example Mean Absolute Error loss for rolling window method on LSTM}
    \label{fig:roll_loss}
  \end{minipage}
  ~
  \begin{minipage}[h]{0.48\textwidth}
    \includegraphics[width=0.98\textwidth]{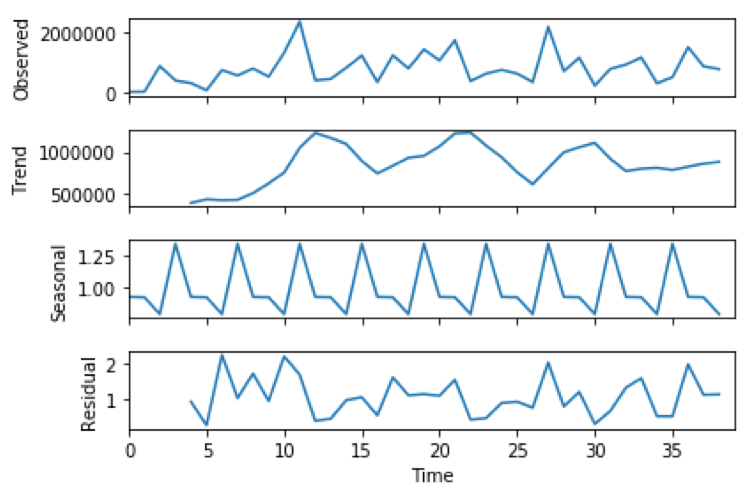}
    \caption{Seasonal decomposition example on one financial datarow}
    \label{fig:stl}
  \end{minipage}
\end{figure}

We present four variants, each cumulatively building upon the previous variant, of our RNN model to show increasing reduction in error.  In all variants, we use a walk-forward split \cite{amita2} wherein validation sets are four steps forward into time from training sets, ensuring no data leakage.  We do this iteratively for windows of size $15$ timesteps within the data, continuously walking forward in time until the end of the training data (i.e., until July 2017); this is referred to as the rolling window process. The window size of $15$ timesteps was chosen empirically. As we move from one window to the next, we use weights obtained from the model trained on data corresponding to the previous window for initialization. An example loss function when using the rolling window process is shown in Figure \ref{fig:roll_loss}; notice that gradual loss is attained as we step through consecutive windows because the model uses prior weights to warm-start rather than fitting from scratch.

\subsubsection{Basic LSTM}
The first model we discuss is our basic RNN model.  All training, validation, and test data are historical financial values that have been smoothed using a logarithmic transformation and de-meaned on training data only.  These pre-processing methods are used throughout due to better experimental results relative to other smoothing transformations.  A single-layer sequence-to-sequence model is fed into a dense layer, using the Adam optimizer on mean absolute error.  The sequence-to-sequence \cite{seq2seq} architecture involves a LSTM encoder (to process revenue and return interal state), and an LSTM decoder (to use the previous time step's actual data and internal LSTM encoder states to generate the next output).  Teacher forcing is used only during training; for inference, we feed in the predicted values for the next timestep as input to the decoder instead of the actual value as would be the case in the teacher forcing method. Next, we apply the inverse smoothing transformation on the decoder's output for last four timesteps (i.e., revenue for the last four quarters) to calculate test error.

\subsubsection{LSTM with Categorical Indicators}
The second model we examine is simply the basic model with additional indicator covariates (i.e., one-hot categorical variables are incorporated in the model).  Specifically, for our three categorical variables (segment, region, and product), we include one-hot encodings so that the hierarchical product information is reflected in the model.
 
\subsubsection{LSTM with Seasonality} 
The third model incorporates seasonal effects in the second model. Specifically, we use multiplicative Seasonal Trend decomposition using Loess (STL) \cite{stl} to calculate trend, seasonal, and residual components.  A sample datarow decomposition is shown in Figure \ref{fig:stl}.  We extract the seasonal component from the relevant datarows, and we use only the product of trend and residual effects (in each quarter, and for each datarow) as inputs to be smoothed and fed to the neural network model. De-seasonalizing the input data along with other aforementioned transformations (logarithmic and de-meaning) helps to make the data more stationary. 

We maintain use of the indicator covariates introduced in the second model.  The only difference now is in the inference step: in addition to decoding and using an inverse smoothing transformation, we must also multiply our predictions obtained from the decoder by the seasonal values calculated for each quarter (timestep) in the previous year.


\subsubsection{LSTM with Curriculum Learning}
\begin{figure}[h]
  \centering
  \begin{minipage}[h]{0.48\textwidth}
    \includegraphics[width=0.98\textwidth]{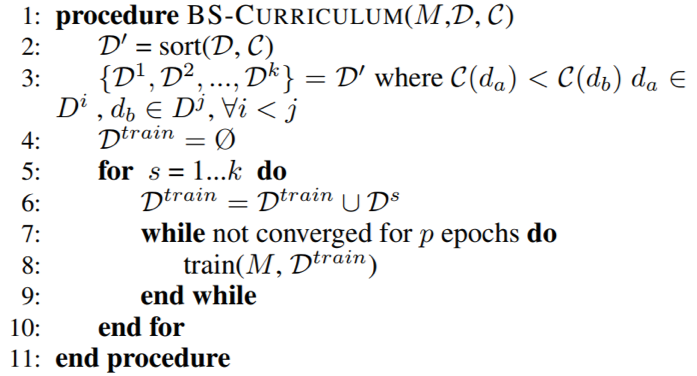}
    \caption{Baby Steps Curriculum Learning Algorithm \cite{babysums,leapfrog}} 
    \label{fig:babysteps}
  \end{minipage}
  ~
  \begin{minipage}[h]{0.48\textwidth}
    \includegraphics[width=0.98\textwidth]{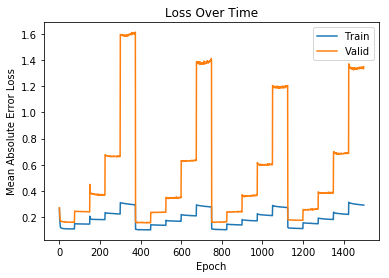}
    \caption{Example Mean Absolute Error loss for curriculum learning with rolling window method on LSTM (ordered from easiest to hardest estimated prediction)}
    \label{fig:curric_loss}
  \end{minipage}
\end{figure}

The fourth model applies curriculum learning to the third model.  We use the pre-calculated seasonal decomposition to determine a useful batch ordering method to feed into our neural net, and then apply the Baby Steps curriculum algorithm \cite{babysums,leapfrog} defined in Figure \ref{fig:babysteps}.


Let us define $D' = sort(D,C)$ for training data $D$ and curriculum sort metric $C$ \cite{babysums}, where our sort metric is specified as the residual trend weighted by segment revenue for each datarow.  That is, we will sort our training data on this new variable we have created, which exists for each datarow. Then, we order our batches such that $\{D^1, D^2, ..., D^k\} = D'$ where $C(d_a) < C(d_b)$ for $d_a \in D^i, d_b \in D^j, \forall i < j$. In words, we train in increasing order of the residual error calculated from the STL decomposition mentioned previously.  Once the $k$ batches are defined, we shuffle the datarows within the batch during training.  Within each iteration of the rolling window process, we continue the warm-start process by iteratively adding one batch at a time to the training data, running each rolling window iteration $p$ times, where $p$ is the number of epochs chosen such that convergence can be reached.  In summary, for each of the $s = 1,...,k$ batches, we run $D^{train} = D^{train} \cup D^s$ until the end of $p$ epochs of each rolling window iteration.  A sample loss function including curriculum learning is shown in Figure \ref{fig:curric_loss}, where we have experimentally chosen $p=75$.

We note that there are several ways to form the batching described above.  In our results, we present batches formed directly from the datarow sort metric calculated as described above, i.e., the datarow-level residual error (using 5 batches, as determined experimentally).  However, we have also found good results when batching by segment (with each batch corresponding to one Microsoft segment), where batches are sorted by revenue-weighted segment-level residual error.  In all cases, we shuffle datarows within each batch when training.  The idea of curriculum learning, as highlighted in Section \ref{sec:intro}, is to train on easier examples to learn harder ones.  As such, we find that running curriculum learning with batches sorted in order of best-fit to least-fit segments yields similar results to those that we find from the datarow-level (uniform-weighting) method we use in this paper.  However, we also experimented with curriculum learning using batches sorted in reverse order: least-fit to best-fit segments.  This resulted in far better accuracies for certain (smaller-revenue) segments, but worse accuracies for all other segments.  Hence, it remains a reasonable option to sort in various ways, and then ensemble results, to tease out best results for different segments.


\subsection{CNN Model: Dilated CNN}
\label{dcnn}

We present three versions of our CNN model to show increased reduction in error across iterations.  We note that we do not explore separating seasonal effects from our DCNN model as a pre-processing step.  While found to be useful in the LSTM model, results are not significantly better for the DCNN when performing seasonal decomposition prior to training the model.  This is likely because the exponential nature of the DCNN layers allows us to capture the seasonality over a long time series, especially since we are using each relevant datarow's full 35-quarter financial history as input to the model (rather than the rolling window method applied for LSTMs).

\subsubsection{Basic Dilated CNN}
In our first Dilated CNN model, we use 1D convolutions in each of 10 dilated convolutional layers (with 6 filters of width 2 per layer).  This connects to an exponential number ($2^{10}$) of input values for the output.  Two fully connected layers are used to obtain a final output: a dense layer of size 128 with ReLU activation, and a dense layer of size 1.  We apply an Adam optimizer on the Mean Absolute Error.  Teacher forcing is done during training only, and predicts the four test quarters of data iteratively, appending each prediction to history for the next timestep's prediction.  Similar to the LSTM model, all historical financial values passed into the DCNN model have been smoothed using a logarithmic transformation and de-meaned on training data only.

\subsubsection{Dilated CNN with Categorical Indicators}
The second DCNN model we examine is simply the above basic model with additional indicator covariates.  Specifically, for our three categorical variables (segment, region, and product), we include one-hot encodings so that the hierarchical product information is reflected in the model. 

\subsubsection{Dilated CNN with Curriculum Learning}
We apply the same mechanism for curriculum learning as explained for the LSTM model, using the residual from seasonal decompositions as a proxy for the difficulty of forecasting each datarow.  We batch by the datarow-level residual error (using 8 batches, as determined experimentally).  The curriculum learning is performed based on the second DCNN model, i.e., including categorical variables, but not using seasonal decomposition for anything aside from the sort order for curriculum learning.

\subsection{Evaluation}
For evaluation purposes, we use the four quarters of data from October 2017 to July 2018 as our test dataset. For certain products, there are only null values available in recent quarters (e.g., if the product is being discontinued) and hence we do not include these products in the test dataset. We use the Mean Absolute Percentage Error (MAPE) as our error evaluation metric. To take into account the inherent randomness involved from weight initialization when training the DNNs, and considering that our data is medium-sized, we run each experiment 30 times to obtain a more robust estimate of the errors. For each datarow, we take the average of the forecasts across runs and across quarters as the final forecast and compare this predicted revenue to the actual observed revenue.  The segment-level forecast is the sum of all (subregion-level and product-level) forecasts falling into that segment.  The world-wide forecast is the sum of forecasts for all datarows.


\section{Results}

We find that both LSTM and DCNN models with curriculum learning out-perform the respective models without curricum learning.  In particular, the Encoder-Decoder LSTM with curriculum learning (including categorical indicators and seasonality effects) yields the lowest error rates, showing a world-wide improvement of 27\%, and a revenue-weighted segment-based improvement of 30\%, over Microsoft's production baseline.  
We further find that curriculum learning models can yield either lower bias or variance for various segments.  

Due to privacy concerns, actual test errors are not displayed.  We instead report relative percentage improvement over the Microsoft baseline in production.  

\subsection{World-wide Error Rates}

World-wide MAPEs for all models are compared in Table \ref{tab:ww_mapes}.  Both LSTM and DCNN models with curriculum learning outperform all variants without curriculum learning by over 10 percentage points.  It is worth noting that even the baseline LSTM model (without curriculum learning) improves upon the Microsoft baseline in production.  We lastly comment on the decrease in world-wide accuracy upon adding seasonality to the LSTM model.  While the world-wide error (MAPE) is higher for this model variant, we see in Table \ref{tab:rnn} that the revenue-weighted segment-level average yields an improvement of 21\% from seasonality over the previous LSTM model variants.  The interpretation here is that seasonality can be more accurately inferred for the few product segments having the largest revenues, and hence the segment-level benefits are outweighed world-wide by the many smaller-revenue datarows that are less accurate (due to more fluctuation in seasonal effects on smaller products).  We suggest that seasonal trend decomposition be used only after careful consideration of the durability of financial seasonality.  In our application, we only present LSTM results including seasonality since we find it beneficial conjointly with curriculum learning; experimentally, our displayed results fare better than the alternative of curriculum learning sans seasonality.

\begin{table}
\centering
\caption{World-wide test error reduction percentages of DNN models over previous Microsoft production baseline.}
\label{tab:ww_mapes}
\begin{tabular}{|l|l|l|l|l|}
\hline
\textbf{Model} &  \textbf{Percent MAPE Improvement}\\
\hline
Basic LSTM &  1.9\% \\
LSTM with Categorical Indicators &  18.2\% \\
LSTM with Seasonality &  -5.1\% \\
LSTM with Curriculum Learning & \textbf{27.0\%} \\
\hline
Basic DCNN &  -0.7\% \\
DCNN with Categorical Indicators &  12.1\% \\
DCNN with Curriculum Learning & \textbf{22.6\%} \\
\hline
\end{tabular}
\end{table}

\begin{figure}[h]
  \centering
  \begin{minipage}[h]{0.48\textwidth}
    \includegraphics[width=0.98\textwidth]{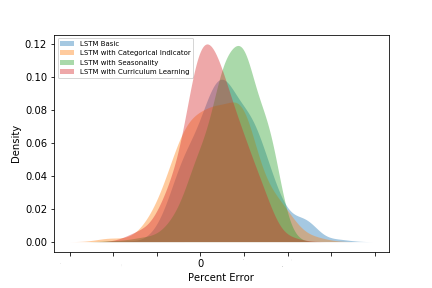}
    \caption{LSTM World-wide Error Density}
    \label{fig:ww_lstm_density}
  \end{minipage}
  ~
  \begin{minipage}[h]{0.48\textwidth}
    \includegraphics[width=0.98\textwidth]{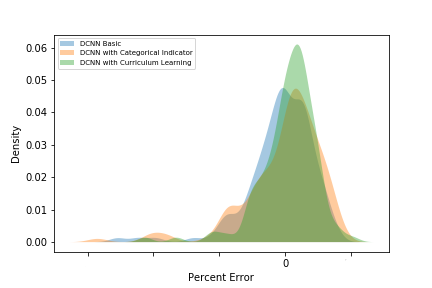}
    \caption{DCNN World-wide Error Density}
    \label{fig:ww_dcnn_density}
  \end{minipage}
\end{figure}

Recall that we ran each model 30 times and took outcome averages to reduce variance.  Density plots are shown for world-wide results in Figures \ref{fig:ww_lstm_density} and \ref{fig:ww_dcnn_density}, which reflect the distribution of calculated (non-absolute) percentage error for each of the LSTM and DCNN models tested, respectively.  These figures allow us to examine the extent to which curriculum learning models are less biased.

\begin{table}
\centering
\caption{LSTM Model Segment-level MAPE reduction percentages (\%) over previous Microsoft production baseline (positive \% corresponds to error reduction).}\label{tab1}
\begin{tabular}{|p{1.5cm}|l|l|l|l|}
\hline
\textbf{Segment} & \textbf{Basic} & \textbf{Model (a) +}  & \textbf{Model(b) +}   & \textbf{Model(c) +}  \\
& \textbf{LSTM} & \textbf{Categorical Indicators} & \textbf{Seasonality} & \textbf{Curriculum Learning} \\
& \textit{\textbf{(Model (a))}}  & \textit{\textbf{(Model (b))}} & \textit{\textbf{(Model (c))}} & \textit{\textbf{(Model(d))}} \\ 
\hline
1 &  25.5 & 22.0 & 53.4 & 70.0 \\
2 &  -47.9 & -34.3 & -23.0 & -0.8 \\
3 &  7.65 & -5.8 & 26.0 & 20.3 \\
4 &  14.2 & 30.3 & 12.0 & 27.4 \\
5 &  -15.4 & -13.2 & -11.8 & -25.9 \\
6 &  -79.2 & -60.3 & -110.1 & -12.4 \\
7 &  34.7 & 30.1 & 31.0 & 11.5 \\
8 &  17.9 & 15.5 & 57.2 & 61.4 \\
\textbf{Revenue-weighted Average} & \textbf{10.3} & \textbf{10.3} & \textbf{21.3} & \textbf{30.0} \\
\hline
\end{tabular}
\label{tab:rnn}
\end{table}

For both density plots, the y-axis denoting density is fully presented.  However, note that the x-axis (expressing percent error) values aside from 0 are excluded for Microsoft privacy reasons.  For both LSTM and DCNN models, we can disclose that the spread of error is bounded by a range of approximately $\pm10$ percentage points.  We claim that applying curriculum learning to our DNN models lessens bias as percent errors are shifted towards zero.



In Figure \ref{fig:ww_lstm_density}, comparisons are displayed among the Basic LSTM (blue), LSTM with Categorical Indicator (orange), LSTM with Seasonality (green), and LSTM with Curriculum Learning (red).  In Figure \ref{fig:ww_dcnn_density}, comparisons are displayed among the Basic DCNN (blue), DCNN with Categorical Indicator (orange), and DCNN with Curriculum Learning (green).  In both cases, we see that the curriculum learning variant (red for LSTM, and green for DCNN) is both closest to being zero-centered and is one of the models with lowest variance.

\subsection{Segment-Level MAPEs}

In Tables \ref{tab:rnn} and \ref{tab:cnn} below, we share the segment-level incremental improvement percentages obtained from the Encoder-Decoder LSTM and DCNN model MAPEs, respectively, as compared to Microsoft's previously-implemented baseline. Due to privacy concerns, the actual names of the segments are not shared since their revenues are considered High Business Impact (HBI) data.

Overall, the revenue-weighted LSTM segment-level MAPEs show a drastic 30\% improvement in MAPE relative to the Microsoft baseline currently used in production (see Table \ref{tab:rnn}). This strong showing is robust both within-segment and world-wide, as seen in Table \ref{tab:ww_mapes}.  Similar gains from curriculum learning are reflected in the DCNN model (see Table \ref{tab:cnn}), showing a 19 percentage point increase from the basic DCNN model to the variant with curriculum learning.

Despite the fact that we cannot discuss segment-specific results in terms of absolute numbers, we comment that it is clear that some segments see significant gains relative to the Microsoft production baseline, whereas others only see modest gains (or slight decreases).  In particular, the addition of curriculum learning with our uniformly-weighted ordering improves results for larger-revenue segments.  However, when using batches based on segment, and using a reverse ordering of segments from hardest to easiest to predict (based on seasonality as we have been doing; results for this variant are not disclosed in this paper), the most improvement is seen in smaller-revenue segments, which otherwise would have been overshadowed by model weights contributing towards larger-revenue segments.  Thus, ensembling these two different sorts is a promising future step.

\begin{table}
\centering
\caption{DCNN Model Segment-level MAPE reduction percentages (\%) over previous Microsoft production baseline (positive \% corresponds to error reduction).}\label{tab1}
\begin{tabular}{|p{1.5cm}|l|l|l|l|}
\hline
\textbf{Segment} & \textbf{Basic} & \textbf{Model (a) + } & \textbf{Model(b) + }  \\
& \textbf{DCNN} & \textbf{Categorical Indicators} & \textbf{Curriculum Learning} \\
& \textit{\textbf{(Model (a))}}  & \textit{\textbf{(Model (b))}} & \textit{\textbf{(Model (c))}}  \\ 
\hline
1 &  24.8 & 44.0 & 34.2 \\
2 &  -0.2 & -19.5 & -19.5 \\
3 &  -8.7 & 28.9 & 39.9 \\
4 &  35.5 & 35.4 & 22.6 \\
5 &  45.4 & 58.4 & 26.8 \\
6 &  -258.2 & -263.2 & -80.5 \\
7 &  27.0 & 28.7 & 29.4 \\
8 &  33.8 & 35.5 & 24.9 \\
\textbf{Revenue-weighted Average} & \textbf{-3.1} & \textbf{4.5} & \textbf{16.2} \\
\hline
\end{tabular}
\label{tab:cnn}
\end{table}

\begin{figure}[h]
  \centering
  \begin{minipage}[h]{0.48\textwidth}
    \includegraphics[width=0.98\textwidth]{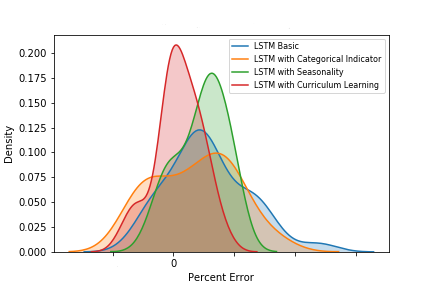}
    \caption{LSTM Larger Segment Error}
    \label{fig:large_lstm_density}
  \end{minipage}
  ~
  \begin{minipage}[h]{0.48\textwidth}
    \includegraphics[width=0.98\textwidth]{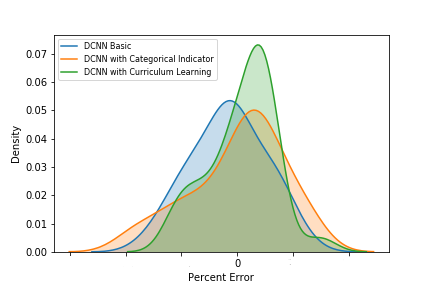}
    \caption{DCNN Larger Segment Error}
    \label{fig:large_dcnn_density}
  \end{minipage}
\\
  \begin{minipage}[h]{0.48\textwidth}
    \includegraphics[width=0.98\textwidth]{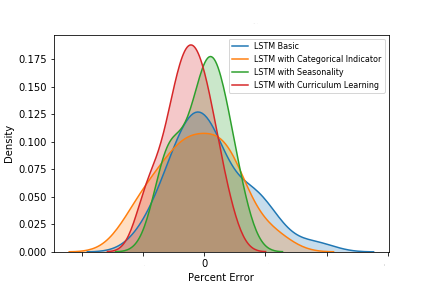}
    \caption{LSTM Smaller Segment Error}
    \label{fig:med_lstm_density}
  \end{minipage}
  ~
  \begin{minipage}[h]{0.48\textwidth}
    \includegraphics[width=0.98\textwidth]{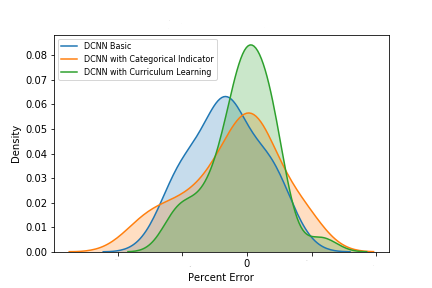}
    \caption{DCNN Smaller Segment Error}
    \label{fig:med_dcnn_density}
  \end{minipage}
\end{figure}

We now turn to examples of segment-specific density plots, which are shown for two specific segments in Figures \ref{fig:large_lstm_density} through \ref{fig:med_dcnn_density}.  For privacy reasons, we cannot disclose the segment names, but we assert that one of the segments has larger revenue, and one of the segments has smaller revenue (amounting to four times less revenue than the larger segment).  In the below plots, we show both segment sizes for which curriculum learning improves results. We use these figures firstly to re-affirm the effect of curriculum learning on bias, but also to comment on across-run variance.
Again, we note that the x-axis (expressing percent error) values aside from 0 are excluded for Microsoft privacy reasons.

We compare the same model variants as in the previous world-wide figures.  Specifically, in Figures \ref{fig:large_lstm_density} and \ref{fig:med_lstm_density}, comparisons are displayed among the Basic LSTM (blue), LSTM with Categorical Indicator (orange), LSTM with Seasonality (green), and LSTM with Curriculum Learning (red).  In Figures \ref{fig:large_dcnn_density} and \ref{fig:med_dcnn_density}, comparisons are displayed among the Basic DCNN (blue), DCNN with Categorical Indicator (orange), and DCNN with Curriculum Learning (green).

We first discuss the larger-revenue segment's density plots (Figure \ref{fig:large_lstm_density} for LSTM models, and Figure \ref{fig:large_dcnn_density} for DCNN models).  We find that the LSTM model with curriculum learning, in particular, improves both bias (towards zero) and variance, as the density spread is decidedly smaller than other model variants.  These effects are less pronounced but similar (especially for lessening variance, but less so for improving bias) for the DCNN model with curriculum learning.

We now discuss the smaller-revenue segment's density plots (Figure \ref{fig:med_lstm_density} for LSTM models, and Figure \ref{fig:med_dcnn_density} for DCNN models).  We find that the DCNN with curriculum learning model, in particular, improves both bias (towards zero) and variance.  These effects are less pronounced but similar (moreso for lessening variance) for the LSTM model with curriculum learning.

Based on the above, we see at the segment level that there can occur a bias and variance trade-off.  Specifically, for the segments wherein curriculum learning improves accuracy, some will see lessened bias (with little change in variance from non-curriculum learning model variants), some will see lessened variance (with little change in bias from non-curriculum learning model variants), and some instances yield better bias and variance.  Hence, we conclude that curriculum learning models on our financial time series for specific segments not only yield more accurate forecasts, but also can achieve relatively low variance and bias.

\section{Discussion}
\label{sec:discussion}

It is clear from our results that there is value in using DNNs on Microsoft's financial time series, and further that curriculum learning is an indispensable tool to improve accuracy of forecasts.  These curriculum learning results are robust both world-wide and at the segment-level.  Further, we see from Figures \ref{fig:ww_lstm_density}  and \ref{fig:ww_dcnn_density} that curriculum learning allows for less bias in errors (robust across both LSTM and DCNN methods), and in certain instances less variance in error at the segment level.

We return to our key methodological takeaways from our work as presented.  First, curriculum learning is a powerful technique for time series data, not just in NLP problems; applying a good sorting metric to neural network batches can improve results drastically.

Second, we contribute much of the efficiency of our DNN methods to transfer learning effects, which are particularly useful for products with a relatively short revenue history.  Here, it is worth noting the importance of data pre-processing.  Executing our DNN methods without regard for missing data yielded worse results than when we subset to training on data with enough historical trends.  Applying the latter model to datarows without sufficient history yielded good results, showing evidence of transfer learning across region, segment, and product.

Lastly, financial data do not need to be extraordinarily large to successfully use neural networks on forecasting.  DNN methods are far more efficient than the Microsoft production baseline that involves ensembling traditional statistical and machine learning methods; it takes a fraction of the time spent to run each DNN model.  While curriculum learning involves sorting, and hence may be unwieldy for very large datasets, it does not significantly impact runtime on the ``medium-sized" Microsoft data, and we are able to create models that do not overfit the data.

Future work includes testing more metrics for curriculum learning, comparing these results to changing sample weights of the hierarchical variables, and ensembling these models for greater accuracy. It would be prudent for future hypertuning packages to include curriculum learning batch metrics and batch sizes as parameters.  We hope to see greater use of DNNs in industry, in particular using curriculum learning on medium-sized datasets. 

\section{Acknowledgements}
We thank Kimyen Nguyen for her generous help with running experiments on security compliant machines considering the sensitivity of the finance data. We also thank Barbara Stortz, Deependra Hamal, and Mindy Yamamoto for their support of this project. The work of A.K. is jointly supported by Microsoft and the National Science Foundation Graduate Research Fellowship under Grant No. DGE – 1656518.   Any opinion, findings, and conclusions or recommendations expressed in this material are those of the authors and do not necessarily reflect the views of the National Science Foundation.

%
%
%
\bibliographystyle{splncs04}
%

\end{document}